\documentclass[sigconf,screen,natbib,nonacm]{acmart}

\def\BibTeX{{\rm B\kern-.05em{\sc i\kern-.025em b}\kern-.08emT\kern-.1667em\lower.7ex\hbox{E}\kern-.125emX}}

\usepackage[paperVersion]{dpupaper}
\newcommand{\projname}{\textsc{CodeSearchNet}\xspace}
\newcommand{\projurl}{\url{https://github.com/github/CodeSearchNet}\xspace}
\newcommand{\task}{\projname\textsc{Challenge}\xspace}
\newcommand{\corpus}{\projname\textsc{Corpus}\xspace}

\begin{document}

\title[\task]{\task\\Evaluating the State of Semantic Code Search}

%
\author{Hamel Husain}
\author{Ho-Hsiang Wu}
\author{Tiferet Gazit}
\email{{hamelsmu,hohsiangwu,tiferet}@github.com}
\affiliation{%
  \institution{GitHub}
}

\author{Miltiadis Allamanis}
\author{Marc Brockschmidt}
\email{{miallama,mabrocks}@microsoft.com}
\affiliation{%
  \institution{Microsoft Research}
}

%
\renewcommand{\shortauthors}{\projname}

%
\begin{abstract}
  \emph{Semantic code search} is the task of retrieving relevant code given a
  natural language query.
  While related to other information retrieval tasks, it requires bridging the gap between
  the language used in code (often abbreviated and highly technical) and
  natural language more suitable to describe vague concepts and ideas.

  To enable evaluation of progress on code search, we are releasing the
  \corpus and are presenting the \task, which consists of 99
  natural language queries with about 4k expert relevance annotations
  of likely results from \corpus.
  The corpus contains about 6 million functions from open-source code spanning
  six programming languages (Go, Java, JavaScript, PHP,
  Python, and Ruby).
  The \corpus also contains automatically generated query-like natural language
  for 2 million functions, obtained from mechanically scraping and preprocessing
  associated function documentation.
  In this article, we describe the methodology used to obtain the corpus and
  expert labels, as well as a number of simple baseline solutions for the task.

  We hope that \task encourages researchers and practitioners to
  study this interesting task further and will host a competition and leaderboard
  to track the progress on the challenge.
  We are also keen on extending \task to more queries and programming
  languages in the future.
\end{abstract}

%
\maketitle

\section{Introduction}
\label{sect:introduction}
The deep learning revolution has fundamentally changed how we approach
perceptive tasks such as image and speech recognition and has shown
substantial successes in working with natural language data.
These have been driven by the co-evolution of large (labelled) datasets,
substantial computational capacity, and a number of advances in
machine learning models.

However, deep learning models still struggle on highly structured data.
One example is \emph{semantic code search}: while search on natural
language documents and even images has made great progress, searching
code is often still unsatisfying.
Standard information retrieval methods do not work well in the code
search domain, as there is often little shared vocabulary between search
terms and results (\eg consider a method called 
\texttt{deserialize\_JSON\_obj\_from\_stream} that may be a correct result for
the query ``read JSON data'').
Even more problematic is that evaluating methods for this task is
extremely hard, as there are no substantial datasets that were created
for this task; instead, the community tries to make do with small datasets
from related contexts (\eg pairing questions on web forums to code
chunks found in answers).

To tackle this problem, we have defined the \task on top of a new
\corpus.
The \corpus was programmatically obtained by scraping open-source repositories
and pairing individual functions with their (processed) documentation
as natural language annotation.
It is large enough (2 million datapoints) to enable training of
high-capacity deep neural models on the task.
We discuss this process in detail in \autoref{sect:traindata}
and also release the data preprocessing pipeline to
encourage further research in this area.

The \task is defined on top of this, providing realistic queries and
expert annotations for likely results.
Concretely, in version 1.0, it consists of 99 natural languages
queries paired with likely results for each of six considered
programming languages (Go, Java, JavaScript, PHP, Python, and Ruby).
Each query/result pair was labeled by a human expert, indicating the
relevance of the result for the query.
We discuss the methodology in detail in \autoref{sect:evaldata}.

Finally, we create a number of baseline methods using a range
of state-of-the-art neural sequence processing techniques (bag of words,
RNNs, CNNs, attentional models) and evaluate them on our datasets.
We discuss these models in \autoref{sect:models} and present
some preliminary results.

\section{The Code Search Corpus}
\label{sect:traindata}
\newcommand\snippet{\ensuremath{\mathbf{c}}}
\newcommand\doc{\ensuremath{\mathbf{d}}}

As it is economically infeasible to create a dataset large enough for
\emph{training} high-capacity models using expert annotations,
we instead create a proxy dataset of lower quality.
For this, we follow other attempts in the literature~\citep{barone2017parallel,gu2018deep,fernandes2018structured,cambronero2019deep}
and pair functions in open-source software with the natural language present
in their respective documentation.
However, to do so requires a number of preprocessing steps and heuristics.
In the following, we discuss some general principles and decisions driven
by in-depth analysis of common error cases.

\paragraph{\corpus Collection}
We collect the corpus from publicly available open-source non-fork GitHub repositories,
using \href{https://libraries.io}{libraries.io} to identify all projects
which are used by at least one other project, and sort them by
``popularity'' as indicated by the number of stars and forks.
Then, we remove any projects that do not have a license or whose license
does not explicitly permit the re-distribution of parts of the project.
We then tokenize all Go, Java, JavaScript, Python, PHP and Ruby
functions (or methods) using \href{https://github.com/tree-sitter/tree-sitter}{TreeSitter} --- GitHub's universal parser ---
and, where available, their respective documentation text using a heuristic regular
expression.

\paragraph{Filtering}
To generate training data for the \task, we first consider
only those functions in the corpus that have documentation associated with
them.
This yields a set of pairs $(\snippet_i, \doc_i)$ where $\snippet_i$
is some function documented by $\doc_i$.
To make the data more realistic proxy for code search tasks, we
then implement a number of preprocessing steps:
\begin{itemize}
 \item Documentation $\doc_i$ is truncated to the first full
  paragraph, to make the length more comparable to search queries
  and remove in-depth discussion of function arguments and return
  values.
 \item Pairs in which $\doc_i$ is shorter than three tokens are
  removed, since we do not expect such comments to be informative.
 \item Functions $\snippet_i$ whose implementation is shorter than
  three lines are removed, these often include unimplemented methods,
  getters, setters, \etc
 \item Functions whose name contains the substring ``test'' are
  removed. Similarly, we remove constructors and standard extension
  methods such as \texttt{\_\_str\_\_} in Python or
  \texttt{toString} in Java.
 \item We remove duplicates from the dataset by identifying (near)
  duplicate functions and only keeping one copy of them (we use the
  methods described in \citet{lopes2017dejavu,allamanis2018adverse}).
  This removes multiple versions of auto-generated code and cases of copy \& pasting.
\end{itemize}
The filtered corpus and the data extraction code are released at
\projurl.

\paragraph{Dataset Statistics}
The resulting dataset contains about 2 million pairs of
function-documentation pairs and about another 4 million functions without
an associated documentation (\autoref{tbl:dsetStats}). We split the dataset in
80-10-10 train/valid/test proportions. We suggest
that users of the dataset employ the same split.

\begin{table}
  \centering
  \begin{tabular}{lrr} \toprule
    & \multicolumn{2}{c}{Number of Functions}\\ \cmidrule{2-3}
    & w/ documentation & All \\ \midrule
Go        & 347\,789 & 726\,768 \\
Java      & 542\,991 & 1\,569\,889 \\
JavaScript& 157\,988 & 1\,857\,835 \\
PHP       & 717\,313 & 977\,821 \\
Python    & 503\,502 & 1\,156\,085 \\
Ruby      & 57\,393 &  164\,048\\ \midrule
All      & 2\,326\,976 & 6\,452\,446\\ \bottomrule
  \end{tabular}
  \caption{Dataset Size Statistics} \label{tbl:dsetStats}
\end{table}

\paragraph{Limitations}
Unsurpsingly, the scraped dataset is quite noisy.
First, documentation is fundamentally different from queries, and hence uses
other forms of language. It is often written at the same time and by the same
author as the documented code, and hence tends to use the same vocabulary,
unlike search queries.
Second, despite our data cleaning efforts we are unable to know the extent to
which each documentation $\doc_i$ accurately describes its associated code
snippet $\snippet_i.$ For example, a number of comments are
outdated with regard to the code that they describe.
Finally, we know that some documentation is written in other languages, whereas
our \task evaluation dataset focuses on English queries.

\section{The Code Search Challenge}
\label{sect:evaldata}
To evaluate on the \task, a method has to return a set of relevant results
from \corpus for each of 99 pre-defined natural language queries.
Note that the task is somewhat simplified from a general
code search task by only allowing full functions/methods as results, and not
arbitrary chunks of code.\footnote{Note that on a sufficiently
large dataset, this is not a significant restriction: more commonly implemented
functionality almost always appears factored out into a function somewhere.}
The \task evaluation dataset consists of the 99 queries with relevance
annotations for a small number of functions from our corpus likely to be
returned.
These annotations were collected from a small set of expert programmers,
but we are looking forward to widening the annotation set going forward.

\paragraph{Query Collection}
To ensure that our query set is representative, we obtained common search queries
from Bing that had high click-through rates to code and combined these with intent
rewrites in StaQC~\citep{yao2018staqc}.
We then manually filtered out queries that were clearly technical keywords (\eg the
exact name of a function such as \code{tf.gather\_nd}) to obtain a set of 99 natural language queries.
While most of the collected queries are generic, some of them are language-specific.

\begin{figure}[t]
    \includegraphics[width=\columnwidth]{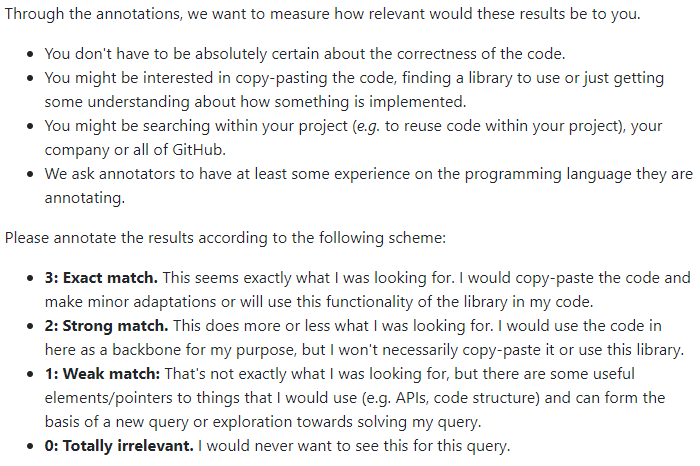}
    \caption{\label{fig:annotator instructions}Instructions provided to annotators.}
\end{figure}

\paragraph{Expert Annotations}
Obviously, we cannot annotate \emph{all} query/function pairs.
To filter this down to a more realistically-sized set, we used our implementations
of baseline methods and ensembled them (see \autoref{sect:models}) to generate 10 candidate results
per query and programming language. Concretely, we used ensembles of all neural models and ElasticSearch
to generate candidate results, merge the suggestions and pick the top 10.
We used a simple web interface for the annotation process.
The web interface firsts shows instructions (see \autoref{fig:annotator instructions})
and then allows the annotator to pick a programming language.
Then, one query/function pair is shown at a time, as shown in \autoref{fig:annotation
interface}.
A link to the origin of the shown function is included, as initial experiments showed
that some annotators found inspecting the context of the code snippet helpful to judge
relevance.
The order of query/code pairs shown to the user is randomized but weakly ordered by the
number of expert annotations already collected.
Annotators are unlikely to see several results for the same query unless they
handle many examples.
By randomizing the order, we aim to allow users to score the relevance of
each pair individually without encouraging comparisons of different results for the same
query.

\begin{figure}[t]
    \includegraphics[width=\columnwidth]{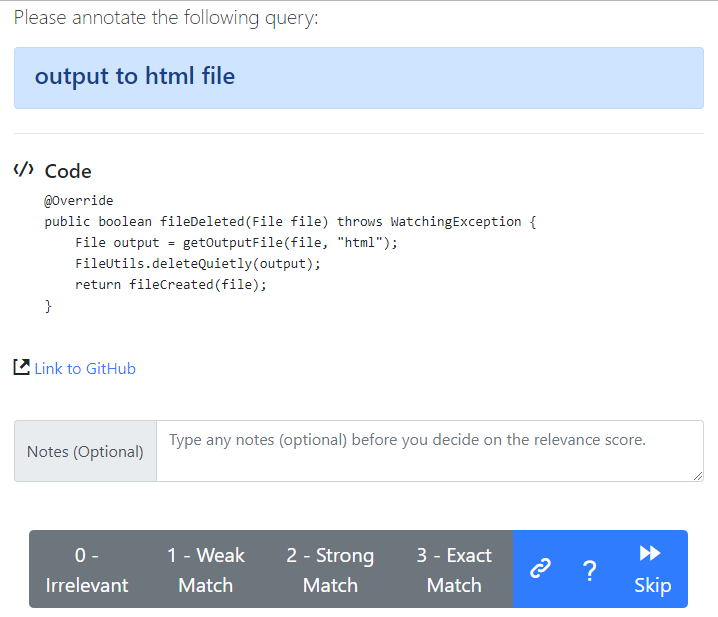}
    \caption{\label{fig:annotation interface}Interface used for relevance annotation.}
\end{figure}

\paragraph{Annotation Statistics}
We collected 4\,026 annotations across six programming languages
and prioritized coverage over multiple annotations
per query-snippet pair.
Our annotators are volunteers with software engineering, data science
and research roles and were asked to only annotate examples for languages
they had significant experience with.
This led to a skewed distribution of annotations w.r.t. the considered
programming languages.

We observed that the obtained relevance scores are distributed differently
for each language (\autoref{tbl:taskStats}).
For example, the relevance scores for Python are evenly distributed across
the four categories while for JavaScript the annotations are skewed towards
lower relevance scores.
There is a number of potential reasons for this, such as
 the quality of the used corpus,
 language-specific interactions with our pre-filtering strategy,
 the queries we collected,
 higher expected relevance standards in the JavaScript community,
 \etc

\begin{table}
  \centering
  \begin{tabular}{lrrrrr} \toprule
    & \multicolumn{4}{c}{Count by Relevance Score} & Total \\ \cmidrule{2-5}
    & 0 & 1 & 2  & 3  & Annotations \\ \midrule
Go        &  62  &  64  & 29  & 11  & 166 \\
Java      & 383  & 178  & 125 & 137  & 823\\
JavaScript& 153  & 52   & 56  & 58   & 319 \\
PHP       & 103  & 77   & 68  & 66   & 314 \\
Python    & 498  & 511  & 537 & 543  & 2\,089\\
Ruby      & 123  & 105  & 53  & 34   & 315\\ \bottomrule
  \end{tabular}
  \caption{Annotation Dataset Statistics} \label{tbl:taskStats}
\end{table}

For the 891 query-code pairs where we have more than one annotation,
we compute the squared Cohen's kappa interannotator agreement to estimate
the quality of the task. The agreement is moderate with Cohen $\kappa=0.47$.
This is somewhat expected given that this task was relatively open-ended,
as we will discuss next.

\paragraph{Qualitative Observations} During the annotation process we
made some observations in discussions with the annotators and
through the notes they provided in the web interface (see Fig. \ref{fig:annotation interface}).
These comments point to some general issues in implementing code search:

\begin{description}
  \item[Code Quality] A subset of the results being returned are functionally
    correct code, but of low quality, even though they originated in
    reasonably popular projects.
    In this context, low quality refers to
     unsatisfactory readability,
     bad security practices,
     known antipatterns
     and potentially slow code.
    Some annotators felt the need to give lower relevance scores to low-quality
    code as they would prefer \emph{not} to see such results.
  \item[Query Ambiguity] Queries are often ambiguous without additional
    context.
    For example, the query ``how to determine if a string is a valid word''
    can have different correct interpretations depending on the
    domain-specific meaning of ``valid word''.
  \item[Library \vs Project Specific] Often a search yields code
    that is very specific to a given project (\eg using internal utility
    functions), whereas other times the code is very general and verbose
    (\eg containing code that could be factored out).
    Which of these is preferable depends on the context of the query,
    which we did not explicitly specify when asking for annotations.
  \item[Context]
    Some results were semantically correct, but not relying on related
    helper functions and thus not self-contained.
    Some annotators were uncertain if such results should be considered relevant.
  \item[Directionality] A common problem in results were functions
    implementing the \emph{inverse} functionality of the query, \eg
    ``convert int to string'' would be answered by \code{stringToInt}.
    This suggests that the baseline models used for pre-filtering have
    trouble with understanding such semantic aspects.
\end{description}

\subsection{Evaluation of Ranking Models}
To track the progress on the \task we have deployed a Weights \& Biases
leaderboard at \url{https://app.wandb.ai/github/codesearchnet/benchmark}.
We hope that this leaderboard will allow the community to better compare
solutions to the code search task.

\paragraph{Metrics} We used normalized discounted cumulative gain (NDCG)
to evaluate each competing method. NDCG is a commonly used metric~\citep{manning2008introduction} in
information retrieval. We compute two variants of NDCG:
 (a) NDCG computed over the subset of functions with human annotations (``Within'')
 (b) NDCG over the whole \corpus (``All'').
We make this distinction as the NDCG score computed over the whole corpus
may not necessarily represent the quality of a search tool, as a new
tool may yield relevant but not-annotated functions.

\section{Baseline CodeSearch models}
\label{sect:models}
We implemented a range of baseline models for the code search task, using
standard techniques from neural sequence processing and web search.

\subsection{Joint Vector Representations for Code Search}
Following earlier work~\cite{gu2018deep,mitra2018introduction}, we use
\emph{joint embeddings} of code and queries to implement a neural search
system.
Our architecture employs one encoder per input (natural or programming)
language and trains them to map inputs into a single, joint vector
space.
Our training objective is to map code and the corresponding language
onto vectors that are near to each other, as we can then implement a
search method by embedding the query and then returning the set of
code snippets that are ``near'' in embedding space.
Although more complex models considering more interactions between
queries and code can perform better~\citep{mitra2018introduction},
generating a single vector per query/snippet allows for efficient
indexing and search.

\begin{figure}[t]
    \includegraphics[width=\columnwidth]{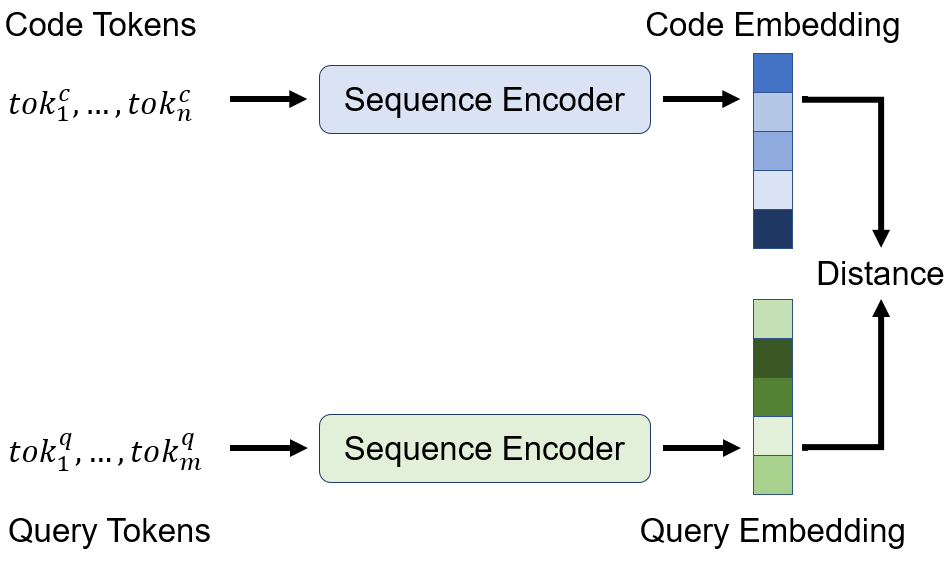}
    \caption{\label{fig:model arch}Model Architecture Overview.}
\end{figure}
To learn these embedding functions, we combine standard sequence encoder
models in the architecture shown in \autoref{fig:model arch}.
First, we preprocess the input sequences according to their semantics:
identifiers appearing in code tokens are split into subtokens (\ie
a variable \texttt{camelCase} yields two subtokens \texttt{camel} and
\texttt{case}), and natural language tokens are split using
byte-pair encoding (BPE)~\citep{gage1994new,sennrich2016neural}.

Then, the token sequences are processed to obtain (contextualized) token
embeddings, using one of the following architectures.
\begin{description}
    \item[Neural Bag of Words] where each (sub)token is
      embedded to a learnable embedding (vector
      representation).
    \item[Bidirectional RNN models] where we employ
      the GRU cell~\citep{cho2014properties} to summarize the input sequence.
    \item[1D Convolutional Neural Network] over the
      input sequence of tokens~\citep{kim2014convolutional}.
    \item[Self-Attention] where multi-head attention~\citep{vaswani2017attention}
      is used to compute representations of each token
      in the sequence.
\end{description}
The token embeddings are then combined into a sequence embedding
using a pooling function, for which we have implemented mean/max-pooling and
an attention-like weighted sum mechanism. For all models, we set the dimensionality
of the embedding space to 128.

During training we are given a set of $N$ pairs $(\snippet_i, \doc_i)$ of code
and natural language descriptions and have instantiated a code encoder $E_c$
and a query encoder $E_q$.
We train by minimizing the loss
\begin{align*}
  -\frac{1}{N}\sum_i
        \log\left(\frac{\exp(E_c(\snippet_i)^\top E_q(\doc_i))}
                       {\sum_ j \exp(E_c(\snippet_j)^\top E_q(\doc_i))}\right),
\end{align*}
\ie maximize the inner product of the code and query encodings
of the pair, while minimizing the inner product between each
$\snippet_i$ and the distractor snippets $\snippet_j$ ($i \ne j$).
Note that we have experimented with other similar objectives (\eg considering
cosine similarity and max-margin approaches) without significant changes in results
on our validation dataset.
The code for the baselines can be found at \projurl.

At test time, we index all functions in \corpus using \href{https://github.com/spotify/annoy}{Annoy}. Annoy offers fast, approximate
nearest neighbor indexing and search. The index includes all functions in the
\corpus, including those that do \emph{not} have an associated documentation
comment. We observed that carefully constructing this index is crucial to achieving
good performance. Specifically, our baseline models were underperforming when we
had a small number of trees in Annoy (trees can be thought as approximate indexes of
the multidimensional space).

\subsection{ElasticSearch Baseline}
In our experiments, we additionally included \href{https://www.elastic.co/products/elasticsearch}{ElasticSearch},
a widely used search engine with the default parameters.
We configured it with an index using two fields for every function in our
dataset: the function name, split into subtokens; and the text of the
entire function.
We use the default ElasticSearch tokenizer.

\subsection{Evaluation}
Following the training/validation/testing data split, we train our baseline
models using our objective from above.
While it does not directly correspond to the real target task of code search,
it has been widely used as a proxy for training similar models~\citep{cambronero2019deep,yao2019coacor}.

\begin{table*}[t]
  \begin{tabular}{llrrrrrrrr} \toprule
  \multicolumn{2}{c}{Encoder} &~& \multicolumn{7}{c}{\corpus (MRR)} \\ \cmidrule{1-2}  \cmidrule{4-10}
  Text & Code && Go & Java & JS & PHP & Python & Ruby & Avg \\ \midrule
  NBoW & NBoW && 0.6409 & 0.5140 & 0.4607 & 0.4835 & 0.5809 & 0.4285 & 0.6167 \\
  1D-CNN & 1D-CNN && 0.6274 & 0.5270 & 0.3523 & 0.5294 & 0.5708 & 0.2450 & 0.6206 \\
  biRNN & biRNN && 0.4524 & 0.2865 & 0.1530 & 0.2512 & 0.3213 & 0.0835 & 0.4262 \\
  SelfAtt & SelfAtt && \textbf{0.6809} & \textbf{0.5866} & 0.4506 & \textbf{0.6011} & \textbf{0.6922} & 0.3651 & \textbf{0.7011} \\
  SelfAtt & NBoW    && 0.6631  & 0.5618  & \textbf{0.4920}  & 0.5083  & 0.6113  & \textbf{0.4574}  & 0.6505 \\
  \bottomrule
  \end{tabular}
  \caption{Mean Reciprocal Rank (MRR) on Test Set of \corpus. This evaluates for our
           training task, where given the documentation comment as a query, the models
           try to rank the correct code snippet highly among 999 distractor snippets.} \label{tbl:mrr eval}
\end{table*}

For testing purposes on \corpus, we fix a set of 999 distractor snippets $\snippet_j$
for each test pair $(\snippet_i, \doc_i)$ and test all trained models.
\autoref{tbl:mrr eval} presents the Mean Reciprocal Rank results on this task.
Overall, we see that the models achieve relatively good performance on this task, with the
self-attention-based model performing best.
This is not unexpected, as the self-attention model has the highest capacity of all
considered models.

We have also run our baselines on \task and show the results in \autoref{tbl:ndcg eval}.
Here, the neural bag of words model performs very well, whereas the stronger neural models
on the training task do less well.
We note that the bag of words model is particularly good at keyword matching,
which seems to be a crucial facility in implementing search methods. This hypothesis
is further validated by the fact that the non-neural ElasticSearch-based baseline performs
competitively among all models we have tested. The NBoW model is the best performing model
among the baselines models, despite being the simplest.
As noted by \citet{cambronero2019deep}, this can be attributed to the fact that the
training data constructed from code documentation is not a good match for the code
search task.

\begin{table*}[t]
  \footnotesize
  \begin{tabular}{llrrrrrrrrrrrrrrrr} \toprule
    \multicolumn{2}{c}{Encoder} && \multicolumn{7}{c}{\task -- NDCG Within} && \multicolumn{7}{c}{\task -- NDCG All} \\ \cmidrule{1-2} \cmidrule{4-10} \cmidrule{12-18}
  Text & Code && Go & Java & JS & PHP & Python & Ruby & Avg && Go & Java  & JS & PHP & Python & Ruby & Avg \\ \midrule
  \multicolumn{2}{c}{ElasticSearch} && 0.307 & 0.257 & 0.318 & 0.338 & 0.406 & 0.395 & 0.337 && 0.186 & 0.190 & 0.204 & 0.199 & 0.256 & 0.197 & 0.205 \\
  NBoW & NBoW && \textbf{0.591} & 0.500 & \textbf{0.556} & 0.536 & 0.582 & \textbf{0.680} & \textbf{0.574} && 0.278 & \textbf{0.355} & \textbf{0.311} & \textbf{0.291} & \textbf{0.448} & \textbf{0.360} & \textbf{0.340} \\
  1D-CNN & 1D-CNN && 0.379 & 0.407 & 0.269 & 0.474 & 0.473 & 0.420 & 0.404 && 0.120 & 0.189 & 0.099 & 0.176 & 0.242 & 0.162 & 0.165 \\
  biRNN & biRNN && 0.112 & 0.165 & 0.066 & 0.148 & 0.193 & 0.185 & 0.145 && 0.030 & 0.056 & 0.017 & 0.042 & 0.070 & 0.060 & 0.046 \\
  SelfAtt & SelfAtt && 0.484 & 0.431 & 0.446 & 0.522 & 0.560 & 0.515 & 0.493 && 0.211 & 0.233 & 0.175 & 0.232 & 0.367 & 0.219 & 0.240 \\
  SelfAtt & NBoW    && 0.550 & \textbf{0.514} & 0.545 & \textbf{0.557} & \textbf{0.583} & 0.650 & 0.566 && \textbf{0.284} & 0.340 & 0.299 & \textbf{0.291} & 0.422 & 0.360 & 0.333 \\
  \bottomrule
  \end{tabular}
  \caption{NDCG Baseline Results on \task. ``Within'' computes the NDCG only on the
          functions within the human-annotated examples. ``All'' computes NDCG over
          all functions in the \corpus.} \label{tbl:ndcg eval}
\end{table*}

\section{Related Work}
\label{sect:related_work}
Applying machine learning to code has been widely considered~\citep{allamanis2018survey}.
A few academic works have looked into related tasks.
First, semantic parsing has received a lot of attention
in the NLP community. Although most approaches are
usually aimed towards creating an executable representation
of a natural language utterance with a domain-specific
language, general-purpose languages have been
recently considered by~\citet{yin2017syntactic,ling2016latent,hashimoto2018retrieve,lin2018nl2bash}.

\citet{iyer2018mapping} generate code from natural language within
the context of existing methods, whereas~\citet{allamanis2016convolutional,alon2018code2seq} consider
the task of summarizing functions to their names. Finally, \citet{fernandes2018structured}
consider the task of predicting the documentation text from
source code.

More related to \projname is prior work in code search
with deep learning. In the last few years there has been
research in this area (\citet{yao2019coacor,gu2018deep,gu2016deep,cambronero2019deep}),
and architectures similar to those discussed previously
have been shown to work to some extent. Recently,
\citet{cambronero2019deep} looked into the same problem that
\projname is concerned with and reached conclusions similar to
those discussed here. In contrast to the aforementioned works, here
we provide a human-annotated dataset of relevance
scores and test a few more neural search architectures
along with a standard information retrieval baseline.

\section{Conclusions \& Open Challenges}
\label{sect:conclusion}

We hope that \projname is a good step towards engaging
with the machine learning, IR and NLP communities
towards developing new machine learning models that
understand source code and natural language. Despite
the fact this report gives emphasis on semantic code search
we look forward to other uses of the presented datasets.
There are still plenty of open challenges in this area.
\begin{itemize}
\item Our ElasticSearch baseline, that performs traditional keyword-based search,
  performs quite well. It has the advantage of being able to
  efficiently use rare terms, which often appear in code.
  Researching neural methods that can efficiently and accurately
  represent rare terms will improve performance.
\item Code semantics such as control and data flow are not
  exploited explicitly by existing methods, and instead search
  methods seem to be mainly operate on identifiers (such as
  variable and function) names.
  How to leverage semantics to improve results remains an
  open problem.
\item Recently, in NLP, pretraining methods such as BERT~\citep{devlin2018bert}
  have found great success. Can similar methods be useful
  for the encoders considered in this work?
\item Our data covers a wide range of general-purpose
  code queries.
  However, anecdotal evidence indicates that queries in specific projects
  are usually more specialized.
  Adapting search methods to such use cases could yield substantial
  performance improvements.
\item Code quality of the searched snippets was
  a recurrent issue with our expert annotators.
  Despite its subjective nature, there seems to be agreement on what
  constitutes very bad code.
  Using code quality as an additional signal that allows for filtering
  of bad results (at least when better results are available) could
  substantially improve satisfaction of search users.
\end{itemize}

\begin{acks}
  We thank Rok Novosel for participating in the CodeSearchNet
  challenge and pinpointing a bug with the indexing that was
  significantly impacting our evaluation results.
\end{acks}

\bibliographystyle{ACM-Reference-Format}
\bibliography{bibliography}

\end{document}